\documentclass[letterpaper, 10 pt, conference]{ieeeconf}  % Comment this line out if you need a4paper

\IEEEoverridecommandlockouts                              % This command is only needed if 
\usepackage{hyperref}
\usepackage{amsmath} % assumes amsmath package installed
\usepackage{amssymb}  % assumes amsmath package installed
\usepackage{cite}
\usepackage{graphics} % for pdf, bitmapped graphics files
\usepackage{graphicx} % for pdf, bitmapped graphics files
\usepackage{subcaption}
\usepackage{float}
\usepackage{tabularx}
\usepackage{multirow}
\usepackage{adjustbox}
\hypersetup{urlcolor=blue, % url
citecolor=black, % citation
linkcolor=black, % table of contents, inner color
colorlinks=true}
	
\title{\LARGE \bf
Colmap-PCD: An Open-source Tool for Fine Image-to-point cloud Registration
}

\author{Chunge Bai$^{1,2}$, Ruijie Fu$^{2}$ and Xiang Gao$^{3}$% <-this % stops a space
\thanks{$^{1}$Chunge Bai is at AgiBot Technology Co. Ltd and Tsinghua University. (email: {\tt\small bcg971004@gmail.com})}%
\thanks{$^{2}$Chunge Bai and Ruijie Fu are in Robotics Institute at Carnegie Mellon University. (email: {\tt\small ruijief@andrew.cmu.edu})}%
\thanks{$^{3}$Xiang Gao is at Robust Robot Technology Co. Ltd. (email: {\tt\small gao.xiang.thu@gmail.com})}%
\thanks{$^{4}$\href{https://github.com/XiaoBaiiiiii/colmap-pcd}{https://github.com/XiaoBaiiiiii/colmap-pcd}}
}

\begin{document}

\maketitle
\thispagestyle{empty}
\pagestyle{empty}

%%%%%%%%%%%%%%%%%%%%%%%%%%%%%%%%%%%%%%%%%%%%%%%%%%%%%%%%%%%%%%%%%%%%%%%%%%%%%%%%
\begin{abstract}
State-of-the-art techniques for monocular camera reconstruction predominantly rely on the Structure from Motion (SfM) pipeline. However, such methods often yield reconstruction outcomes that lack crucial scale information, and over time, accumulation of images leads to inevitable drift issues. In contrast, mapping methods based on LiDAR scans are popular in large-scale urban scene reconstruction due to their precise distance measurements, a capability fundamentally absent in visual-based approaches. Researchers have made attempts to utilize concurrent LiDAR and camera measurements in pursuit of precise scaling and color details within mapping outcomes. However, the outcomes are subject to extrinsic calibration and time synchronization precision. In this paper, we propose a novel cost-effective reconstruction pipeline that utilizes a pre-established LiDAR map as a fixed constraint to effectively address the inherent scale challenges present in monocular camera reconstruction. To our knowledge, our method is the first to register images onto the point cloud map without requiring synchronous capture of camera and LiDAR data, granting us the flexibility to manage reconstruction detail levels across various areas of interest. To facilitate further research in this domain, we have released Colmap-PCD${^{3}}$, an open-source tool leveraging the Colmap algorithm, that enables precise fine-scale registration of images to the point cloud map.

\end{abstract}
%%%%%%%%%%%%%%%%%%%%%%%%%%%%%%%%%%%%%%%%%%%%%%%%%%%%%%%%%%%%%%%%%%%%%%%%%%%%%%%%

\section{INTRODUCTION}
\begin{figure}[!t]
	\centering
	\includegraphics[width=0.45\textwidth]{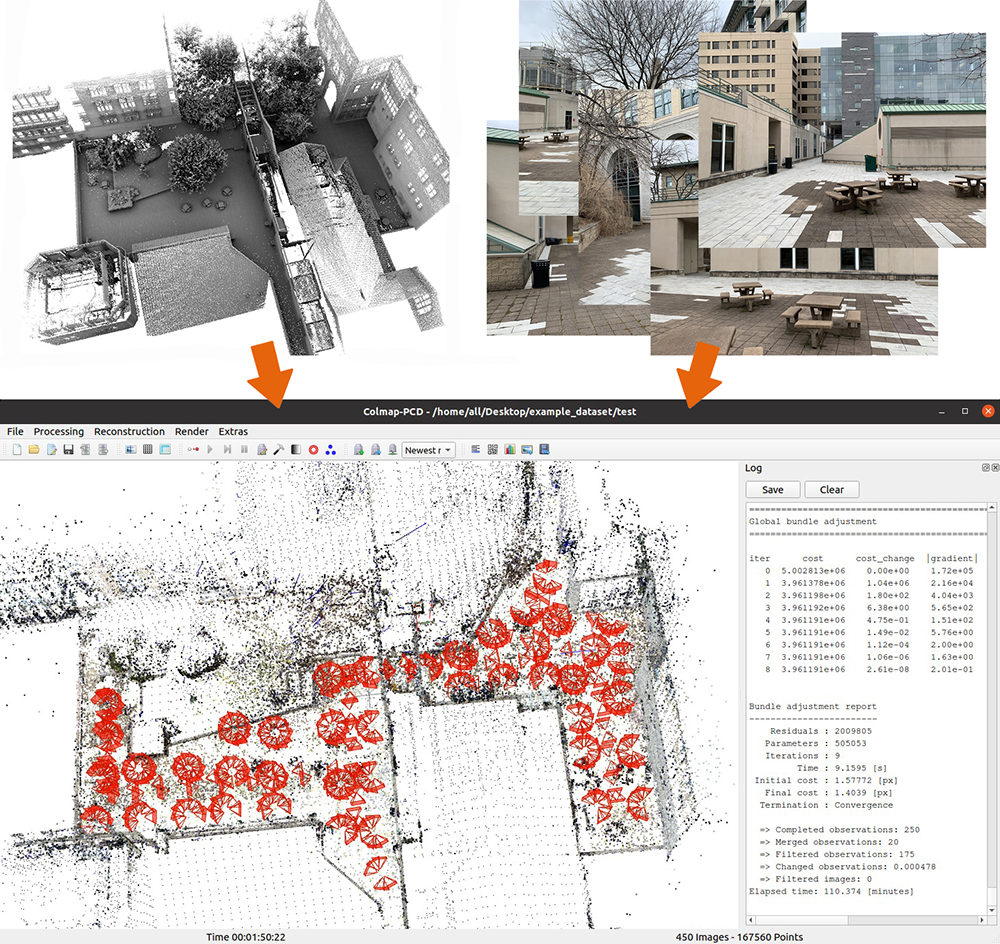}
	\caption{The mapping user interface of Colmap-PCD. Input data include the point cloud and images.}
	\label{fig:ui}
\vspace{-0.5cm}
\end{figure}

Image localization is crucial in the field of robotics and augmented reality, where achieving precise image localization is a critical component for creating realistic scene reconstructions. Typically, image localization and 3D feature point reconstruction are built on the well-established Structure-from-Motion (SfM) pipeline \cite{schoenberger2016sfm}. However, despite the increasing popularity and continuous advancements in SfM algorithms, persistent challenges such as accuracy and scale ambiguity remain to be solved, particularly in the context of large-scale reconstruction. As the number of images increases, the inevitable occurrence of drifts poses a significant challenge. In contrast, LiDAR mapping leverages its ability to capture 3D data in the form of point clouds, providing inherent and highly accurate scaling information within the 3D space. This attribute makes LiDAR mapping particularly well-suited for tasks involving large-scale urban scene reconstruction. Therefore, adopting a collaborative mapping strategy that harnesses the strengths of both LiDAR and visual sensors holds great promise for achieving scale-accurate and robust 3D scene reconstruction.

However, even when collecting images and LiDAR scans simultaneously, the quality of the mapping results remains significantly influenced by the precision of extrinsic calibration and time synchronization. Furthermore, the acquisition of image data often occurs at a fixed frequency, leading to an inefficient allocation of computational resources since not every region necessitates the same level of reconstruction accuracy. Consequently, if it becomes possible to separately acquire LiDAR data and images, it would offer substantial flexibility in adjusting image capture frequencies. This approach would allow higher sampling rates in regions abundant in detail, while reducing sampling in areas with fewer details, thus aligning with specific reconstruction requirements. In addition, this approach proves to be cost-effective by eliminating the necessity for an exceptionally expensive LiDAR system. Users can opt to rent a LiDAR unit for the preconstruction of point cloud and subsequently cease its usage, depending on more affordable and easily accessible cameras for image localization and reconstruction. Moreover, collecting data asynchronously also benefits agent collaboration. After the initial LiDAR mapping is done, we can follow up with image capture using multiple camera models, which are then registered to the same LiDAR point cloud map at different timestamps. This streamlined collaboration reduces the hardware needs and eases the time synchronization demands on data acquisition equipment.

Despite the aforementioned advantages, registering LiDAR point clouds with images can be a formidable task when data is not collected synchronously. Achieving precise 2D-3D correspondences between images and their reconstructed objects is crucial for accurate image localization. One promising approach involves establishing correspondences between images and LiDAR planes, which can be considered as substitute 3D features of the reconstructed objects derived from the LiDAR point cloud map. For this purpose, we choose to employ Colmap\cite{schoenberger2016sfm, schoenberger2016mvs}, a robust software tool based on the classical Structure-from-Motion (SfM) pipeline. Building upon this foundation, we augment the pipeline by incorporating a pre-established LiDAR map as a fixed constraint within the factor graph optimization process. This inclusion helps ensure that the reconstructed scale of the images closely matches that of the LiDAR map, enhancing the accuracy of the registration process. Our method is evaluated using self-collected datasets, and the results clearly demonstrate its superior performance over the standard Colmap approach. More specifically, our approach produces a reconstruction result with an accurate scale, which falls beyond the capability of the original Colmap.

In summary, this paper presents three significant contributions: 1) Introducing Colmap-PCD, an image-to-point cloud registration pipeline that refines image localization using LiDAR maps. It produces accurate localization and reconstruction with scale information, without requiring synchronous LiDAR and camera data collection. 2) Demonstrating the efficacy of Colmap-PCD through comprehensive testing on self-collected datasets. 3) Publishing an open-source tool and datasets to the research community. The user interface is shown in Fig.~\ref{fig:ui}.

\section{RELATED WORK}
We will introduce the commonly used features in LiDAR and visual mapping and the mathematical algorithms frequently employed in 3D reconstruction.

\textbf{Metric Features}: For LiDAR mapping, point-to-plane and point-to-line features \cite{faster-lio, loam} are usually applied to the 3D data accepted by the multi-line LiDAR. Meanwhile, point types such as ground points \cite{legoloam} and pole points \cite{YuePan2021MULLSVL} can be used to segment the point cloud and assist with feature matching. In visual reconstruction, feature descriptors such as SIFT \cite{sift}, SURF \cite{bay2006surf} and ORB \cite{orb} are commonly employed to identify the same features between images. Direct methods like SVO \cite{forster2014svo} and LSD-SLAM \cite{engel2014lsd} leverage the pixel-level grayscale invariance to recognize the motion direction of pixels.
% These approaches rely on the extraction and matching of salient features from images to establish correspondences and estimate the camera's motion. 
% SIFT \cite{sift} features are invariant to scale, rotation, and partially invariant to illumination changes and affine transformations. 
% The innovation of SIFT lies in its keypoint detection using the Difference of Gaussian (DoG) approach and the generation of a feature descriptor based on the local image gradient. The advantages of SIFT include its robustness to various image transformations, making it suitable for matching images under different conditions. 
% SURF \cite{bay2006surf} is another notable feature extraction technique, which is inspired by SIFT but aims to provide a faster and more efficient alternative. 
% ORB \cite{orb} algorithm combines the FAST \cite{fast} keypoint detector with the BRIEF descriptor while incorporating rotation invariance. 
% The advantages of ORB are its computational efficiency and lower memory requirements compared to SIFT and SURF, making it suitable for real-time applications and embedded systems. One of the most influential works in this area is ORB-SLAM \cite{ORB-SLAM22016, ORB-SLAM32020}. The authors introduced a SLAM system based on the efficient extraction and matching of ORB features, resulting in a robust, accurate, and computationally efficient solution suitable for real-time applications.

\textbf{Photogrammetry}: 
%Results of feature matching can be used to estimate the relative camera pose transformation and generate corresponding 3D points through triangulation. 
For a monocular camera, epipolar geometry \cite{MultipleVG} uses an essential matrix or a fundamental matrix to describe the pose transformation relationship between frames. 
%If the feature points in the scene lie on the same plane, the homography matrix can be used to describe the projection relationship between the plane where the feature points are located and the image.
For stereo cameras or RGB-D cameras, methods from the iterative closest point (ICP) \cite{icp1992} series are used to estimate the transformation relationship between two sets of 3D points. When one set comprises 3D points and the other consists of 2D points, the Perspective-n-Point (PnP) \cite{p3p,upnp} method serves as a technique to solve for the motion of the 3D-to-2D point correspondence. EPnP \cite{EPnP} is usually used to avoid the influence of outliers. P4P \cite{p4p} and sampling-based approaches \cite{ArnoldIrschara2009FromSP} can be used for uncalibrated cameras.

\textbf{Factor graph-based approaches}: 
%Using factor graphs to solve for the optimal state is applicable across a wide range of SLAM problems. 
ISAM \cite{kaess2008isam} employs an incremental update mechanism that efficiently updates the nodes and edges affected by new measurements when they are added, avoiding the need to reconstruct the entire graph. In visual SLAM, the reprojection error is a common cost in various algorithms like OKVIS series\cite{okvis2013,okvis2015}. In LiDAR-based SLAM, point-to-plane or point-to-point distance errors are commonly used cost functions. In LiDAR-Vision SLAM, V-LOAM \cite{zhang2015vloam} and LIMO \cite{graeter2018limo} combine point cloud registration from LiDAR with feature tracking from images within a pose-graph optimization framework, and the fusion of these complementary modalities enhances the system's robustness and accuracy.
% VINS-Mono \cite{vinsmono2018} uses a tightly-coupled, nonlinear optimization-based method to obtain high-accuracy visual-inertial odometry by fusing pre-integrated IMU measurements and feature observations.
\begin{figure*}[t!]
\vspace{0.2cm}
	\centering
	\includegraphics[width=1 \textwidth]{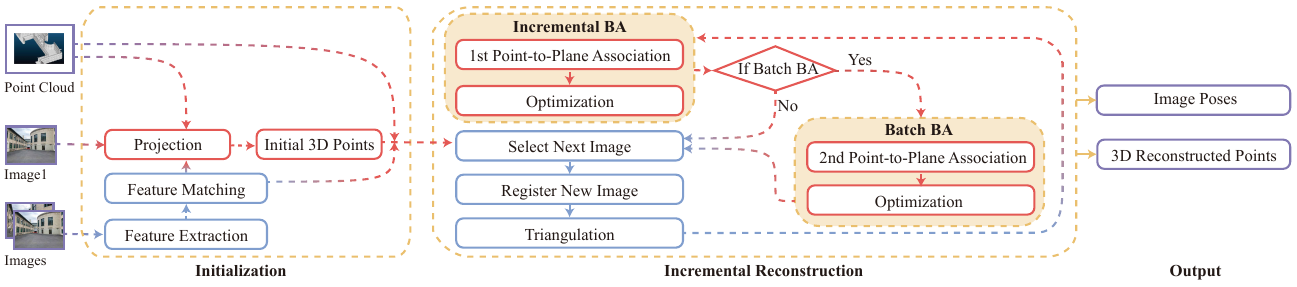}
	\caption{The pipeline of Colmap-PCD. The process in the blue box belongs to the original Colmap, and the process in the red box is the LiDAR-related process.}
	\label{fig:pipeline}
\vspace{-0.7cm}
\end{figure*}
\section{REVIEW of COLMAP}
Here is an introductory overview of Colmap. For more detailed information, please refer to the original paper \cite{schoenberger2016sfm}.

\textbf{Correspondence Search}: The correspondence search process consists of feature extraction and feature matching, identifying matching 2D feature points across various input images to determine potential scene overlap. To enhance the reliability of image matching, Colmap chooses the most credible transformation model by calculating the number of inliers under different transformation conditions.

\textbf{Initialization}: Select an appropriate image pair for the initial reconstruction, ensuring it shares sufficient common observations with multiple images. Then, utilize triangulation to generate the initial set of 3D reconstruction points.
% allowing subsequent images to be registered onto the correct structure.

\textbf{Select and Register Next Image}: Select an image that has more common visible points with the current model and a more uniform distribution pattern as the next image for registration. Register the image to the model by PnP.

\textbf{Triangulation}: Perform triangulation on the new image to integrate the new scene into the existing 3D structure. Due to noise, Colmap treats the feature tracks of a single 3D point as a group of measurements, which can be paired. Use the Direct Linear Transform (DLT) method \cite{MultipleVG} to calculate the 3D point of the paired measurements. Reliable triangulation results are obtained through recursive sampling using RANSAC and adjusting the sampling area. 

\textbf{Incremental Bundle Adjustment}: After each triangulation, local bundle adjustment (BA) is conducted to adjust the parameters and 3D points' positions of the newly registered image and other registered images that share more common observations. 

\textbf{Global Bundle Adjustment}: Global bundle adjustment is performed to compute the optimal 3D model and parameters for all registered images. To save time, global bundle adjustment (BA) will be conducted after the model has grown by a certain degree compared to its volume at the previous global optimization.

% Re-Triangulation (RT) will be performed before BA and after BA to improve the completeness of the reconstruction. After BA, some points that are tracked failure before due to inaccurate poses might be successfully tracked. This step attempts to merge tracks and provide increased redundancy for the next BA step. BA, RT, and filtering outliers are performed in an iterative optimization until the number of filtered observations and post-BA RT points diminishes.

\section{PROBLEM DEFINITION}

Define $\mathcal{I} = \{{I_i}\ |\ i=1 ...N_I\}$ as the input images waiting for pose estimations. And define ${\mathcal{L}=\{(l_m, n_m)\ |\ m=1...N_L\}}$ as the input LiDAR point set, ${l_m \in \mathbb{R}^3}$ represents the 3D coordinate of the LiDAR point, ${n_m \in \mathbb{R}^3}$ represents the unit normal vector of the plane where the LiDAR point is located. Note that the unit normal vector should be supplied, and it can be computed by fitting a surface on neighboring points. 

Denote $\mathcal{F}^i=\{x_j^i\ |\ j=1...N_{F_i}\}$ as the image features extracted from the image ${I_i}$, ${x_j^i \in \mathbb{R}^2}$ is the pixel position.

Denote ${\mathcal{S}=\{\mathbf{X_m}\ |\ m=1...N_{S}\}}$ represents the 3D points that are reconstructed in the model, ${\mathbf{X_m} \in \mathbb{R}^3}$ is the coordinate of the 3D point in the world coordinate system, and ${\mathcal{E}(\mathbf{X_m})}$ represents the track 2D features of ${\mathbf{X_m}}$, ${\mathcal{E}(\mathbf{X_m})=\{{x_s^i\ |\ x_s^i \in \mathcal{F}^i,\ I_i \in \mathcal{I}_m}\}}$. ${x_s^i}$ is the 2D feature on the image ${I_i}$ which should be the projection pixel of the ${\mathbf{X_m}}$ and ${\mathcal{I}_m}$ denotes the image set that can observe ${\mathbf{X_m}}$.

The variable states we want to acquire are the image poses ${\mathcal{T}=\{(\mathbf{R_i}, \mathbf{t_i})\ |\ i=1...N_I\}}$ which are defined on a high-dimensional manifold ${\mathrm{SO}(3) \times \mathbb{R}^3}$, representing the transformations from the world to the camera, where ${\mathbf{R_i}\in \mathrm{SO}(3)}$ is the rotation matrix of the image ${I_i}$, ${\mathbf{t_i}\in \mathbb{R}^3}$ is the translation vector of the image ${I_i}$ from the camera center point to the world coordinate system origin. 

\textit{Problem: }Given the input image set ${\mathcal{I}}$, the LiDAR map ${\mathcal{L}}$, the initial pose guess ${\mathbf{(R_1, t_1)}}$ of the initial image ${I_1}$, and the 2D features ${\mathcal{F}}$ extracted from images, find the optimal image poses ${\mathcal{T}}$, and the 3D points ${\mathcal{S}}$ generated by reconstruction.

\section{COLMAP-PCD ALGORITHM}

Colmap-PCD matches 3D points reconstructed from visual images with the planes extracted from the LiDAR point cloud map, aiming to simultaneously minimize the reprojection error and the distances between the 3D points and the associated LiDAR planes. This section will elaborate on the Colmap-PCD algorithm concretely, and the whole pipeline is shown in Fig.~\ref{fig:pipeline}.

\subsection{Projection}
\begin{figure}[!htp]
	\centering
	\includegraphics[width=0.45\textwidth]{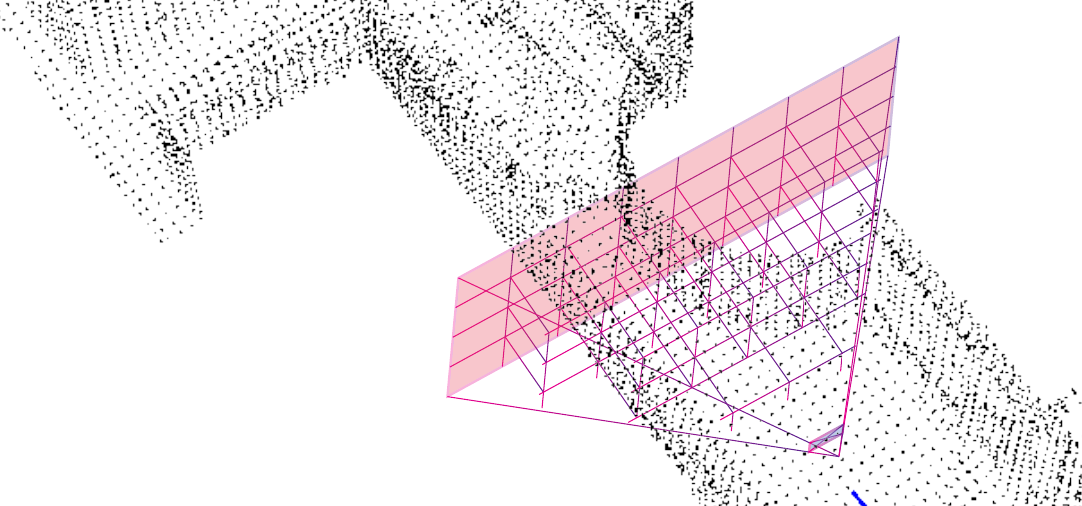}
	\caption{Project LiDAR points in the rectangular pyramid to the camera imaging plane. The vertex of the pyramid represents the camera center.}
	\label{fig:projection}
\vspace{-0.7cm}
\end{figure}

\begin{figure}[!htp]
\vspace{0.2cm}
	\centering
	\includegraphics[width=0.45\textwidth]{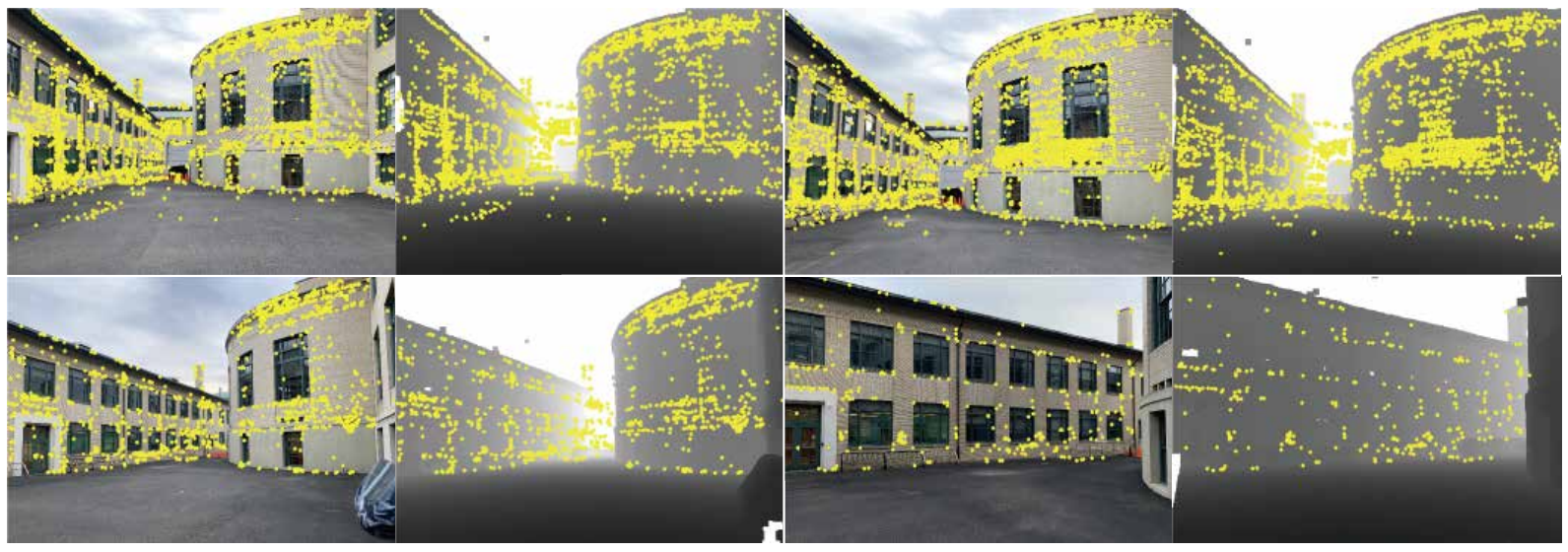}
	\caption{Original images and depth images. The yellow points represent the 2D feature points extracted from the image. In the depth image, the darker the color, the closer the distance.}
	\label{fig:depth-img}
\vspace{-0.7cm}
\end{figure}

The projection method is used in the initialization step and 1st point-to-plane association step, which means to project LiDAR point cloud to a known camera imaging plane ${I_n}$ as shown in Fig.~\ref{fig:projection}. 
Segment the LiDAR point cloud into fixed-size voxels. From the camera center, four lines are emitted outward through the four corners of the normalized image, forming a quadrangular pyramid. The height of the quadrangular pyramid is an adjustable parameter. Traverse the voxels in the quadrangular pyramid, and ${\mathcal{L}_n}$ denotes all LiDAR points in the voxels. For each LiDAR point ${l_e}$ in ${\mathcal{L}_n}$, transform its coordinate from the world coordinate system to the camera's coordinate system, then project it onto the imaging plane:
\begin{equation}\label{eq:LiDAR_point_proj}
	\begin{aligned}
		& l_e^{'} = \mathbf{R}_n \cdot l_e + \mathbf{t}_n \\
		& x_e^p[0] = f_x \cdot l_e^{'}[0]/l_e^{'}[2] + c_x \\
		& x_e^p[1] = f_y \cdot l_e^{'}[1]/l_e^{'}[2] + c_y
	\end{aligned}
\end{equation}
where ${l_e^{'}}$ is the ${l_e}$'s coordinate on the camera's coordinate system, ${x_e^p}$ is the 2D pixel on the image ${I_n}$ that has been projected by the ${l_e}$.
Pay attention to following the projection principle. After projection, the projection pixel of the farthest LiDAR points doesn't change, while the projection pixel of the nearby LiDAR points will be scaled up according to the distance between the LiDAR points and the image, and the closer the distance, the greater the magnification factor. Whether the scale factor of the projection is appropriate can be observed by outputting the depth image as Fig.~\ref{fig:depth-img}. After being extended, ${x_e^p}$ usually denotes more than one pixel. If ${x_e^p}$ equals to a 2D feature point ${x \in \mathcal{F}^n}$ of the ${I_n}$, ${l_e}$ will be seen as a candidate corresponding LiDAR point of ${x}$. Because ${x}$ may accept more than one LiDAR point's projection, the nearest LiDAR point among the candidates will be chosen as the corresponding LiDAR point. This way, we can get the corresponding LiDAR points of some 2D features in ${\mathcal{F}^n}$.

\subsection{Point-to-Plane Association}
Point-to-plane association means corresponding a 3D point to a LiDAR point. The 3D point should fit the plane to which the LiDAR point belongs as closely as possible, with the distance between the 3D point and the plane minimized. Here, we introduce two ways to associate a 3D point to a LiDAR plane: the depth projection method and the nearest neighbor (NN) searching method. We first introduce the depth projection method used in the 1st point-to-plane association step. Assume that 3D point ${\mathbf{X}_k}$ can be observed by the new registered image ${I_n}$ and its latest track 2D feature is ${x_a^n}$, ${{x_a^n} \in \mathcal{E}(\mathbf{X}_k)}$, if the image ${I_m}$ has enough common feature matches with ${I_n}$ and ${x_b^m \in \mathcal{E}(\mathbf{X}_k)}$ (the matching threshold can be set manually), ${I_m}$ will be done projection to search for the corresponding LiDAR point to ${x_b^m}$.
In this way, ${\mathbf{X_k}}$ may correspond to more than one LiDAR point because it has more than one track 2D feature.
For each LiDAR point, compute the angle between the line connecting the LiDAR point and the projection 2D feature point and the normal vector of the Lidar plane, and choose the LiDAR point with the smallest angle as the corresponding LiDAR point. The LiDAR plane of the corresponding LiDAR point is seen as the corresponding LiDAR plane to the ${\mathbf{X}_k}$.

The nearest neighbor searching method will be used in all point-to-plane association procedures. 
Given a 3D point ${\mathbf{X}_k}$, kd-tree is used to search a LiDAR point that has the shortest Euclidean distance to ${\mathbf{X}_k}$. 
The search distance decreases linearly as the number of optimizations for ${\mathbf{X}_k}$ increases. The plane the corresponding LiDAR point is on is seen as the corresponding LiDAR plane to the ${\mathbf{X}_k}$.

\subsection{Initialization}
The initial image requires an approximately known camera pose to initialize the reconstruction process. This pose denotes the camera center's position within the point cloud. The initial image and its pose are set manually on the user interface. The pose doesn't need to be precise and can be acquired by the point cloud tool like Cloud Compare. 
For the initial image ${I_1}$, parts of 2D features in ${\mathcal{F}^1}$ can get the corresponding LiDAR points by making a projection, and the distance between the 2D feature and the LiDAR plane is seen as the depth information of the 2D feature. With the depth information of the feature points, the initial set of 3D points in the current camera coordinate system is obtained. The initial rough 3D model ${\mathcal{X}_1}$ is obtained by converting these 3D points from the camera coordinate system to the LiDAR coordinate system. Although ${\mathcal{X}_1}$ is not accurate enough, it will gradually converge to the correct states as the number of images increases.
This way promises the mapping scale during initialization is approximately accurate. 
% The second image is selected through the original Colmap process by computing the feature matching. 
% Then, because the image ${I_1}$ and ${I_2}$ have common features, the second image can be registered to ${\mathcal{X}_1}$.

\subsection{Bundle Adjustment}

\begin{figure}[ht]
\vspace{-0.3cm}
	\centering
	\includegraphics[width=0.45\textwidth]{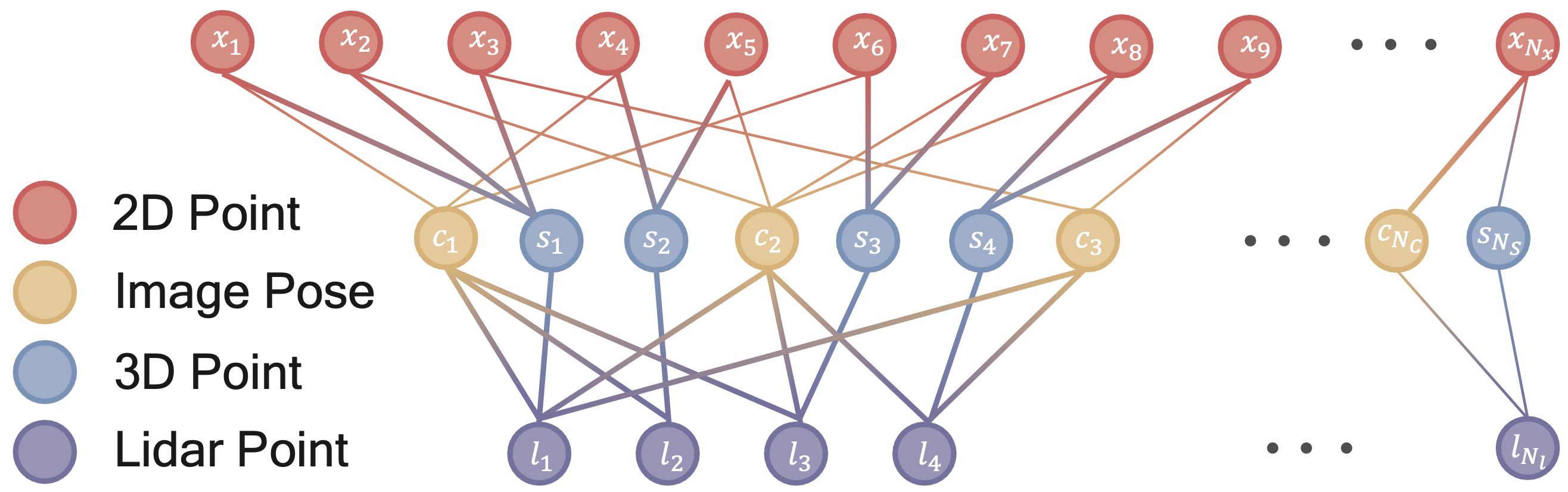}
	\caption{Factor graph for bundle adjustment.}
	\label{fig:factor_graph}
\vspace{-0.3cm}
\end{figure}

There are three types of bundle adjustment: incremental bundle adjustment, batch bundle adjustment, and whole map bundle adjustment. After a new image's triangulation, incremental BA will be performed to mitigate accumulated errors. Batch BA will be performed after a few registrations to save time. After the incremental reconstruction, a 3D model consisting of a series of camera poses and 3D points is built. Whole Map BA could be applied to optimize the whole model further. Bundle adjustment means adding 2D features, image poses, 3D points, and LiDAR points to the factor graph as factors for optimization, as shown in Fig.~\ref{fig:factor_graph}.

2D features and LiDAR points are fixed factors; image poses and 3D points might be constant or variable, depending on the type of bundle adjustment. The reprojection process is added to the factor graph as the 2D point-to-3D point link and the 2D point-to-image pose link. The reprojection error ${\mathbf{E}_r}$ is defined as:
\begin{equation}\label{eq:repro_error}
	\begin{aligned}
		&\mathbf{X}_k^c = \mathbf{R}_i \times \mathbf{X}_k + \mathbf{t}_i \\
		&\mathbf{e}_{r}^{ik} = \sqrt{(\mathbf{X}_k^c[0]/\mathbf{X}_k^c[2] - x_d^i[0])^2 + (\mathbf{X}_k^c[1]/\mathbf{X}_k^c[2] - x_d^i[1])^2} \\
		&\mathbf{E}_r = \sum{\mathbf{e}_{r}^{ik}},\ k=1...N_{\mathbf{s}},\ x_d^i \in \mathcal{E}(\mathbf{X}_k)\\
	\end{aligned}
\end{equation}
where ${\mathbf{X}_k}$ is a 3D point, ${\mathbf{R}_i, \mathbf{t}_i}$ is the camera pose of the image ${I_i}$, ${\mathbf{X}_k^c}$ is the coordinate of the ${\mathbf{X}_k}$ on the camera coordinate system. ${x_d^i}$ belongs to the ${\mathbf{X}_k}$'s track 2D features, ${\mathbf{e}_{r}^{ik}}$ is the reprojection error of the ${\mathbf{X}_k}$ on the image ${I_i}$'s normalized plane. ${\mathbf{E}_r}$ denotes the sum of all reprojection errors. ${N_s}$ is the number of the 3D points.

For the 3D point ${\mathbf{X_k}}$ that has found corresponding LiDAR point ${\mathbf{l_q}}$, ${\mathbf{l_q}}$ is added to the factor graph as a constraint to the ${\mathbf{X_k}}$. ${l_q}$ will be identified into three types. According the searching method, ${l_q}$ will be regarded as projected ${l_q^p}$ or nearest neighbor (NN) ${l_q^n}$. By the direction of the normal vector ${n_q}$, if ${n_q}$ is towards the z-axis, ${l_q}$ will be judged as ground ${l_q^g}$, different point type has different optimized weight ${w^p, w^n, w^g}$. The distance between the ${\mathbf{X_k}}$ and the plane of the ${\mathbf{l_q}}$ is used as the distance error ${\mathbf{E}_d}$ for optimization: 
\begin{equation}\label{eq:corr_error}
	\begin{aligned}
		\mathbf{E}_d = \sum {w \times \| \mathbf{X}_k \cdot n_q -n_q  \cdot l_q\|}, \ k = 1...N_s
	\end{aligned}
\end{equation}
where ${\mathbf{X}_k}$ is a 3D point, ${(l_q, n_q)}$ is ${\mathbf{X}_k}$'s corresponding LiDAR point, ${w}$ is the weight of the ${l_q}$. ${N_s}$ is the total number of the 3D points that have found corresponding LiDAR points.

Bundle adjustment solving for the optimal image poses and 3D point states by minimizing the sum of ${\mathbf{E}_r}$ and ${\mathbf{E}_d}$.

\subsubsection{Incremental Bundle Adjustment}

Incremental BA is performed on a closely connected set of images (have more common features with the new registered image ${I_n}$) because a new registration only affects the model locally. 
For a registered 3D point ${X}$, whether its state will be adjusted in this step depending on the number of its registered track 2D features. If the number reaches the threshold, ${X}$ will be set to be constant; otherwise, it will be seen as a variable state. For variable 3D points, 1st point-to-plane association is applied to search their corresponding planes. In 1st point-to-plane association procedure, 3D points are divided into two groups. The first group owns the 3D points tracked by ${I_n}$, and their corresponding planes will be searched by the depth projection method. Other 3D points are in the second group, and their corresponding planes will be acquired by the nearest neighbor searching method. The search radius is decreased as the number of searches increases. After 1st point-to-plane association, the poses of the image set closely connected to ${I_n}$ will be set to variables, and other image poses will be set to be constant. Then, variable image poses and 3D points will be optimized by the fixed constraints constructed by the constant image poses, 2D features, and LiDAR planes.

\subsubsection{Batch Bundle Adjustment}

The poses of images in the same local area are correlated. Some images may have been registered on the map very early but are in the same area as ${I_n}$. The same scene is observed from different viewpoints, and unified optimization can reduce the drift of pose estimation and improve local coordination. 

\begin{figure}[ht]
\vspace{-0.4cm}
	\centering
	\includegraphics[width=0.3\textwidth]{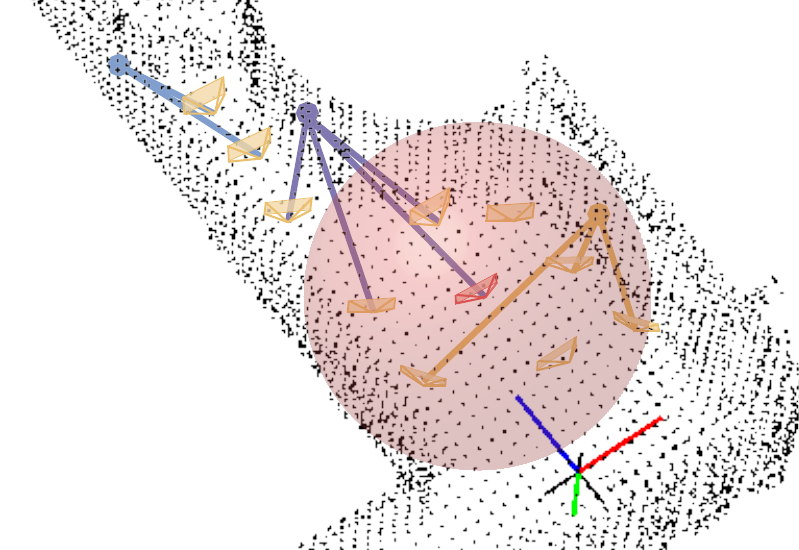}
	\caption{Add images in the sphere to the factor graph for optimization.}
	\label{fig:sphere}
\vspace{-0.3cm}
\end{figure}
As Fig.~\ref{fig:sphere}, we use the camera center of ${I_n}$ as the sphere center, making a sphere with a fixed radius (an adjustable parameter). Images in the sphere are added to the factor graph as variable factors. The blue 3D point isn't observed by the images in the sphere, so it is set to be constant. The purple 3D point and the yellow 3D point can be observed by images in the sphere, so they are added to the factor graph as variable factors. The nearest neighbor searching method is used to find the corresponding planes to these variable 3D points, called the 2nd point-to-plane association. In this BA step, the poses of images inside the sphere and the positions of 3D points that these images can observe are variable.

\subsubsection{Whole Map Bundle Adjustment}

For each 3D point, use NN searching to find its corresponding LiDAR plane. All image poses and 3D points in the model are variable.

\subsection{Incremental reconstruction based on known poses}

If the rough cameras' poses are already known by the measurements of some sensors like GPS, the incremental reconstruction can be started with the known poses.
A pose file written with the image poses can be uploaded through the user interface. The known pose will be used as the prior value to start the triangulation.

\section{EXPERIMENTS}
\begin{figure}[h]
\vspace{-0.2cm}
	\centering
	\begin{subfigure}[b]{0.15\textwidth}
		\includegraphics[width=\linewidth]{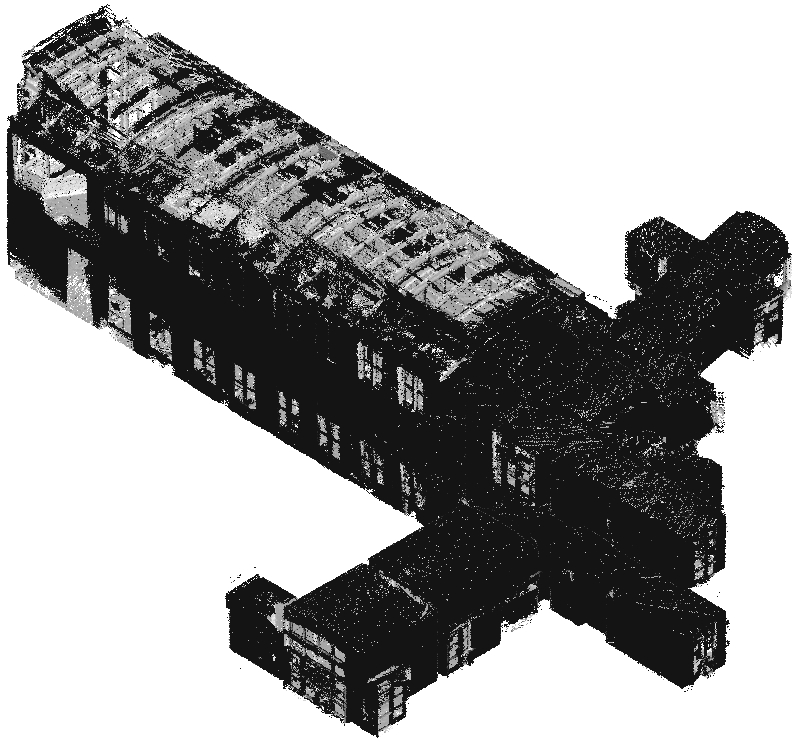}
		\caption{indoor dataset.}
		\label{fig:subfig1}
	\end{subfigure}
	\begin{subfigure}[b]{0.15\textwidth}
		\includegraphics[width=\linewidth]{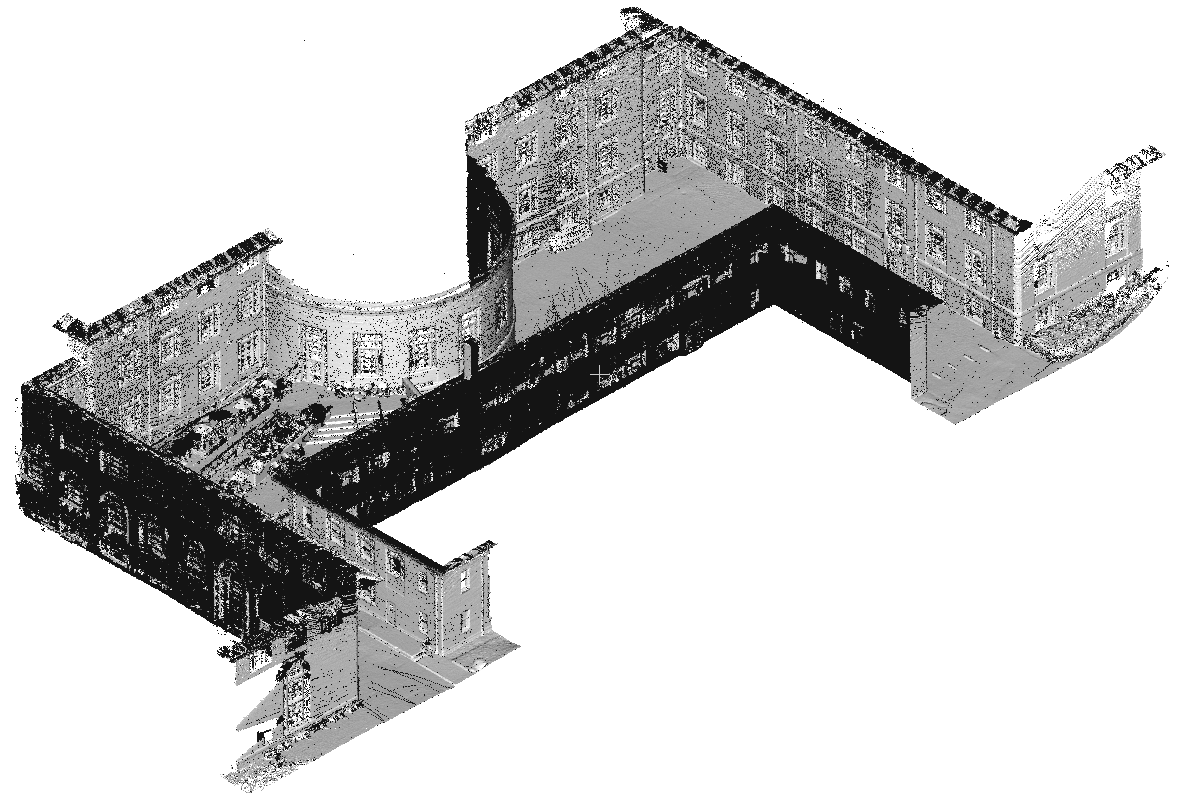}
		\caption{outdoor dataset1.}
		\label{fig:subfig2}
	\end{subfigure}
	\begin{subfigure}[b]{0.15\textwidth}
		\includegraphics[width=\linewidth]{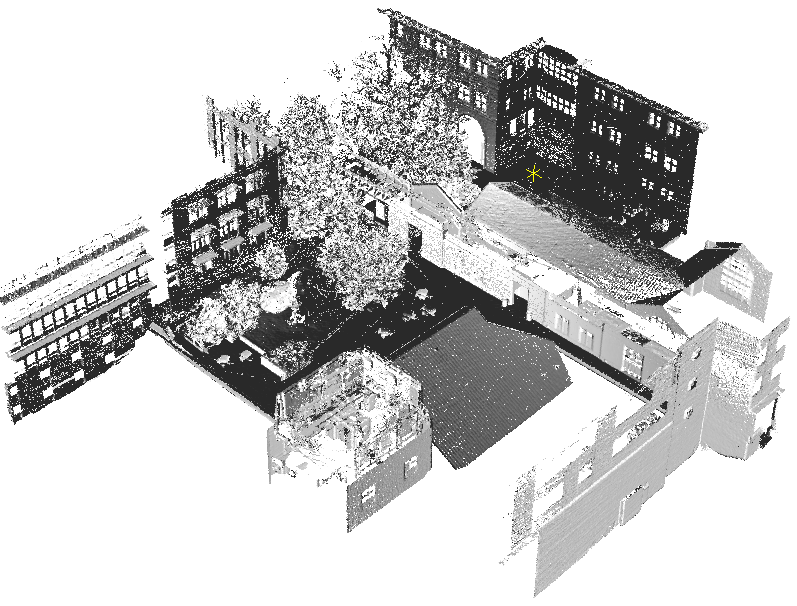}
		\caption{outdoor dataset2.}
		\label{fig:subfig3}
	\end{subfigure}
	\caption{Self-collected point cloud map.}
	\label{fig:datasets}
\vspace{-0.6cm}
\end{figure}

We verify the effectiveness and scale accuracy of Colmap-PCD. 
We collected three datasets at the campus as shown in Fig.~\ref{fig:datasets}. 
Each dataset has a pre-built LiDAR point cloud map, 450 images, and corresponding camera intrinsics.

\subsection{Localization Results}
\vspace{-0.3cm}
\begin{figure}[ht]
	\centering
	\begin{subfigure}{0.15\textwidth}
			\includegraphics[width=\linewidth]{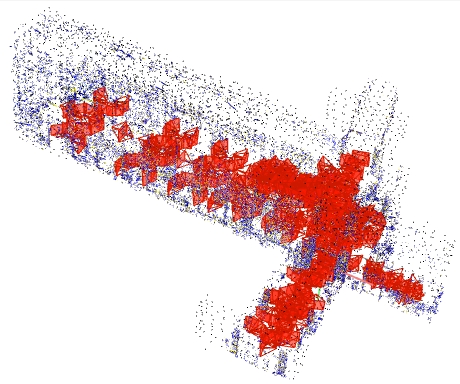}
	\end{subfigure}
	\begin{subfigure}{0.15\textwidth}
			\includegraphics[width=\linewidth]{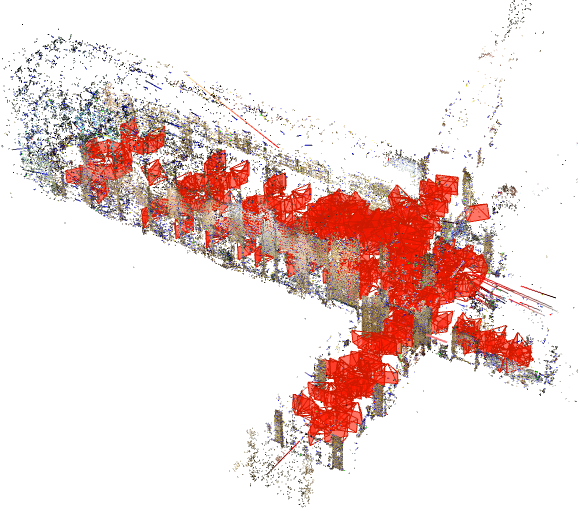}
	\end{subfigure}
	\begin{subfigure}{0.15\textwidth}
		\includegraphics[width=\linewidth]{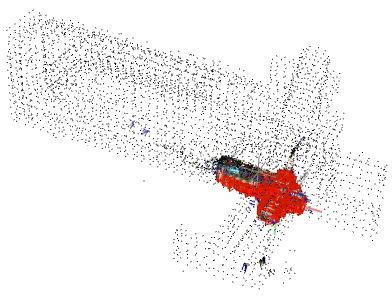}
	\end{subfigure}
	% Second row
	\centering
	\begin{subfigure}{0.15\textwidth}
			\includegraphics[width=\linewidth]{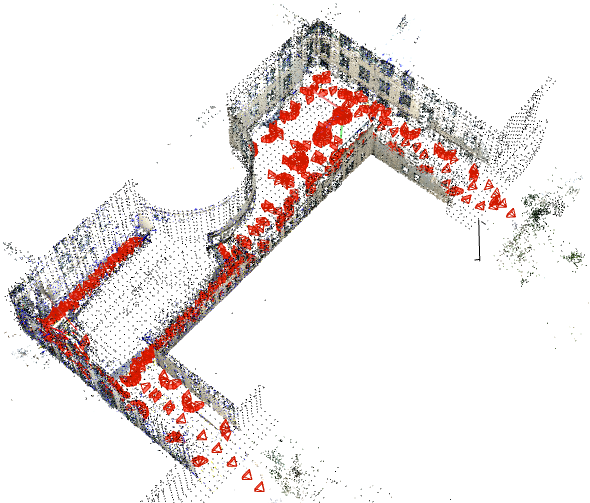}
	\end{subfigure}
	\begin{subfigure}{0.15\textwidth}
			\includegraphics[width=\linewidth]{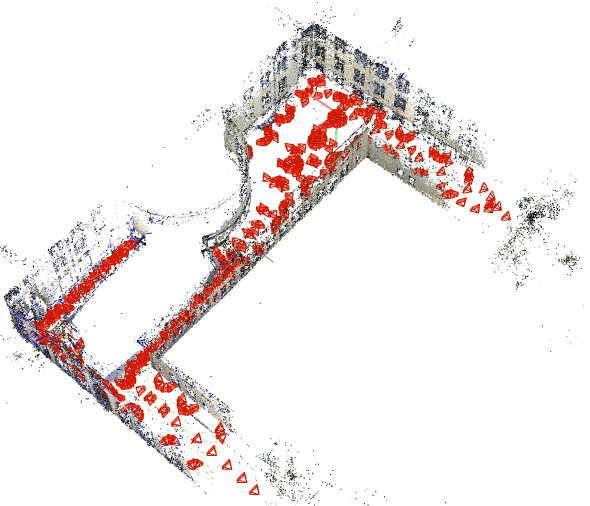}
	\end{subfigure}
	\begin{subfigure}{0.15\textwidth}
		\includegraphics[width=\linewidth]{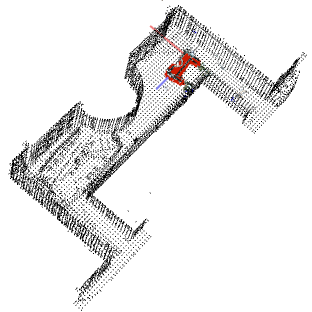}
	\end{subfigure}
	% Third row
	\centering
	\begin{subfigure}{0.15\textwidth}
		\includegraphics[width=\linewidth]{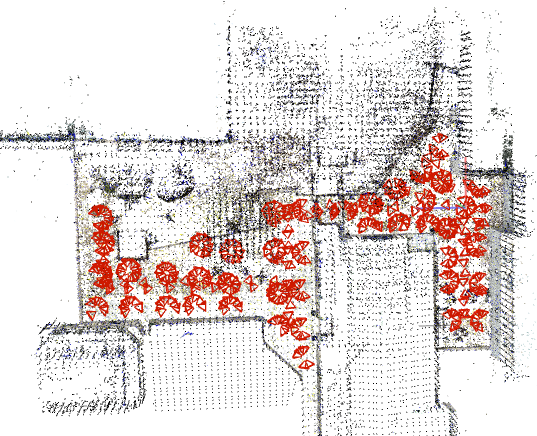}
	\end{subfigure}
	\begin{subfigure}{0.15\textwidth}
			\includegraphics[width=\linewidth]{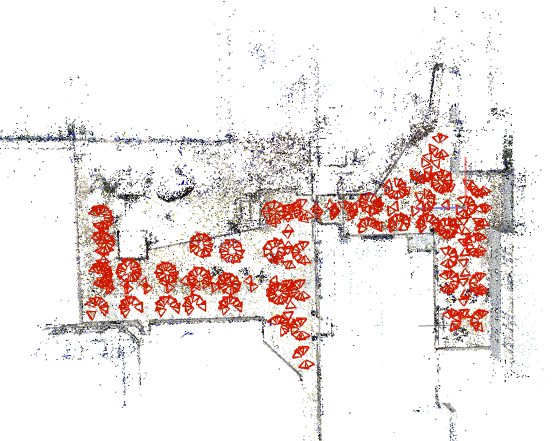}
	\end{subfigure}
	\begin{subfigure}{0.15\textwidth}
		\includegraphics[width=\linewidth]{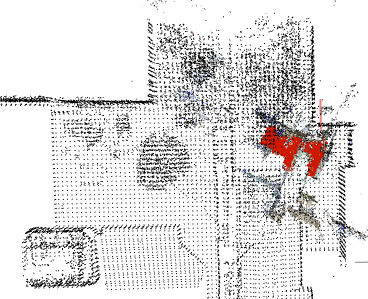}
	\end{subfigure}
	
	\caption{Reconstruction result of Colmap-PCD and Colmap.}
	\label{fig:reconstruction_result}
\vspace{-0.3cm}
\end{figure}
Fig.~\ref{fig:reconstruction_result} shows the reconstruction results of Colmap-PCD and the original Colmap. 
The left and middle images display the reconstructed results of Colmap-PCD. 
The left image comprises both the LiDAR map and the 3D reconstructed points, while the middle image only shows the 3D reconstructed points. 
The right image shows the reconstructed result of Colmap, including the LiDAR map and the 3D reconstructed points.
The red pyramids represent camera poses.
Apparently, the reconstruction result aligns almost perfectly with the LiDAR-generated map, indicating that the LiDAR point cloud map provides a positive constraint in the image localization process.
However, the mapping scale of the original Colmap significantly differs from the real environment.

\begin{figure}[h]
\vspace{0.3cm}
	\centering
	\begin{subfigure}[b]{0.25\textwidth}
		\includegraphics[width=\linewidth]{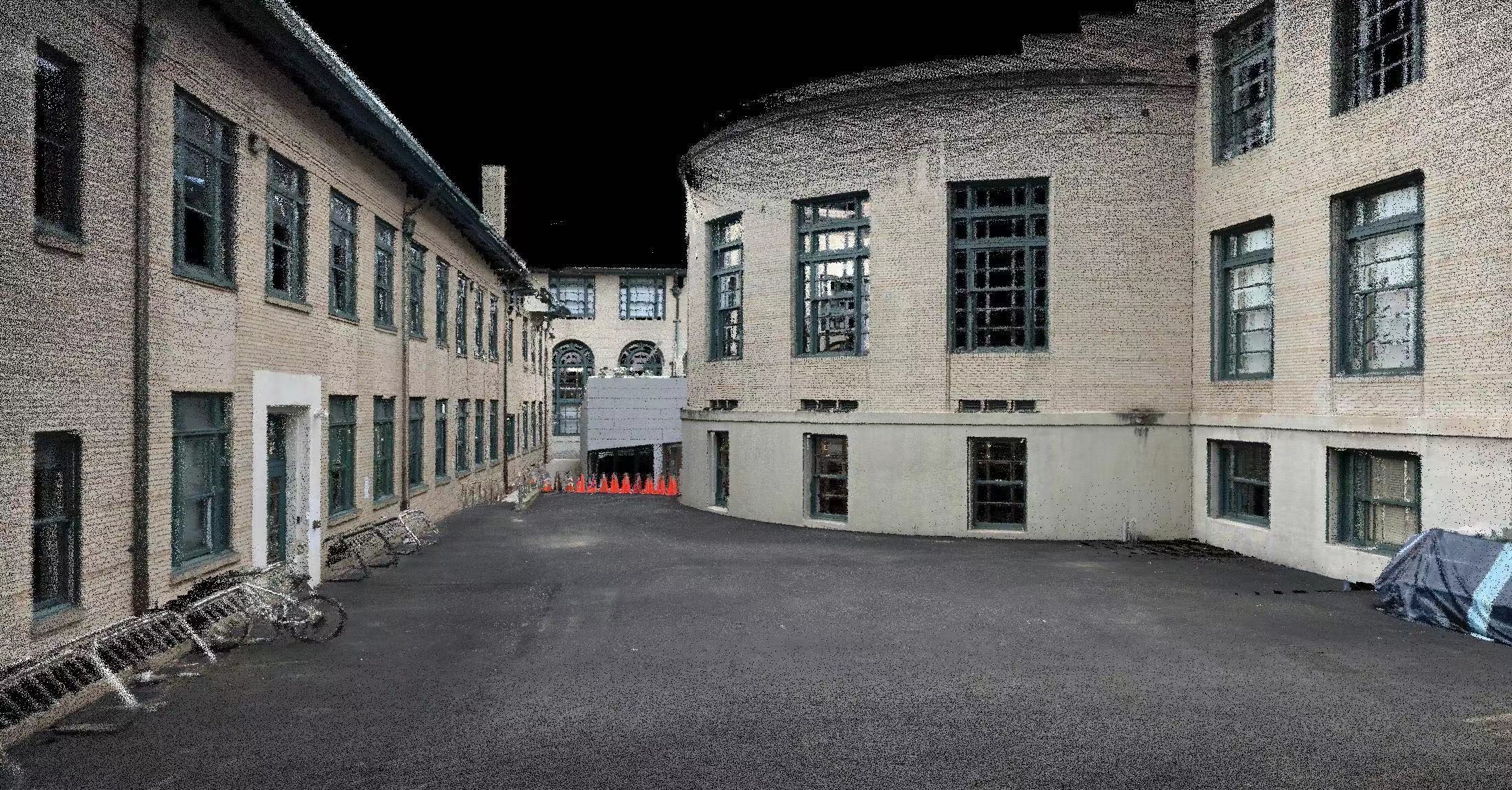}
		\caption{Outdoor dataset1}
		\label{fig:subfig1}
	\end{subfigure}
	\begin{subfigure}[b]{0.22\textwidth}
		\includegraphics[width=\linewidth]{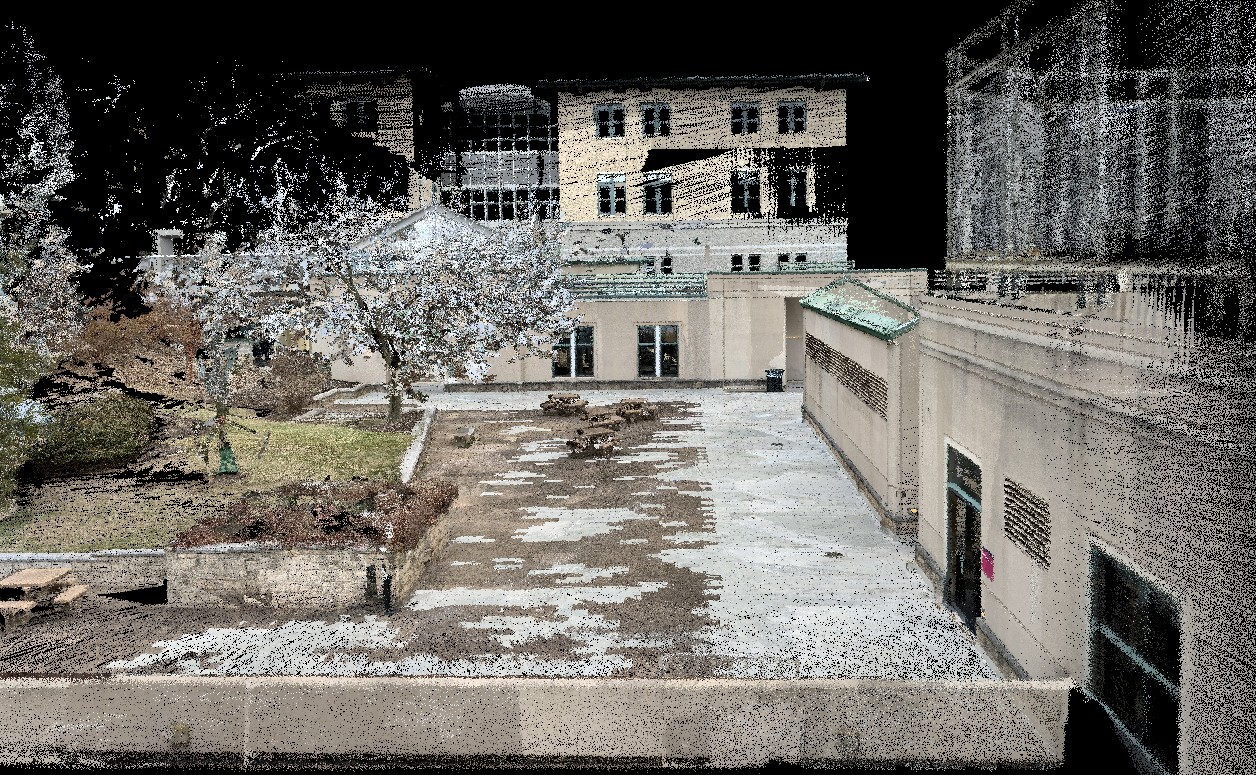}
		\caption{Outdoor dataset2}
		\label{fig:subfig2}
	\end{subfigure}
	\caption{Colored point cloud.}
	\label{fig:projection_result}
\vspace{-0.5cm}
\end{figure}
Fig.~\ref{fig:projection_result} displays the back-projection result from images onto the LiDAR point cloud, which is obviously accurate.
The result demonstrates that Colmap-PCD can achieve the necessary accuracy for image localization.

\subsection{Influence of LiDAR Points on Image Localization}
\begin{figure}[!htp]
	\begin{subfigure}{\textwidth}
		\includegraphics[width=0.35\textwidth]{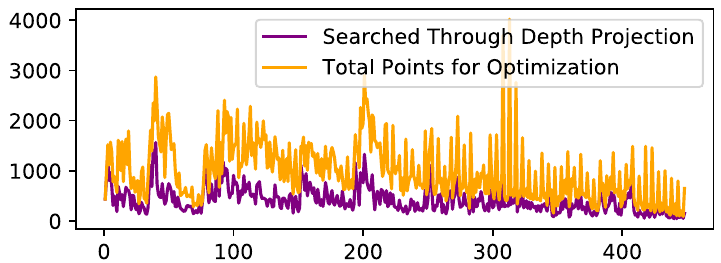}
	\end{subfigure}
	\begin{subfigure}[b]{0.23\textwidth}
		\includegraphics[width=\linewidth]{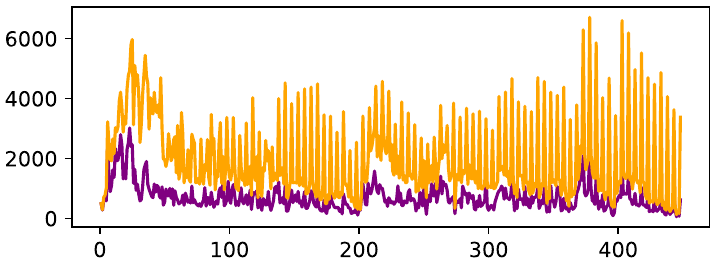}
		\label{fig:subfig1}
	\end{subfigure}
	\begin{subfigure}[b]{0.23\textwidth}
		\includegraphics[width=\linewidth]{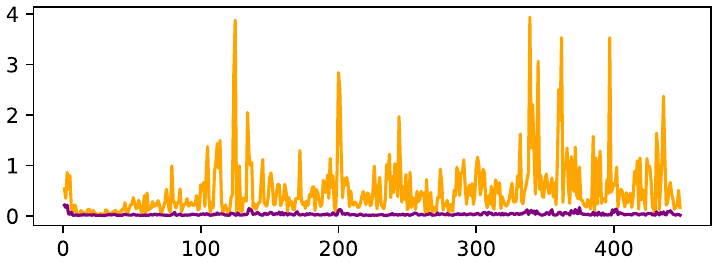}
		\label{fig:subfig2}
	\end{subfigure}
\vspace{-0.3cm}
	\caption{Number of planes used in incremental bundle adjustment (indoor, outdoor1, outdoor2).}
	\label{fig:num_LiDAR}
\vspace{-0.3cm}
\end{figure}
The first experiment examines the number of LiDAR planes that can be used to benefit the pose estimation process. Planes are searched through point-to-plane association.
As shown in Fig.~\ref{fig:num_LiDAR}, the x-axis represents the number of solved image poses, and the y-axis represents the count of LiDAR planes utilized in a single iteration. 
The purple line represents that the matching between the LiDAR plane and the 3D point is achieved through projection, and the yellow line represents the total count of LiDAR planes that are successfully associated through both nearest neighbor searching and projection.
It can be observed that a considerable number of LiDAR planes assist in image localization.

\begin{figure}[!htp]
\vspace{0.1cm}
	\begin{subfigure}{\textwidth}
		\includegraphics[width=0.35\textwidth]{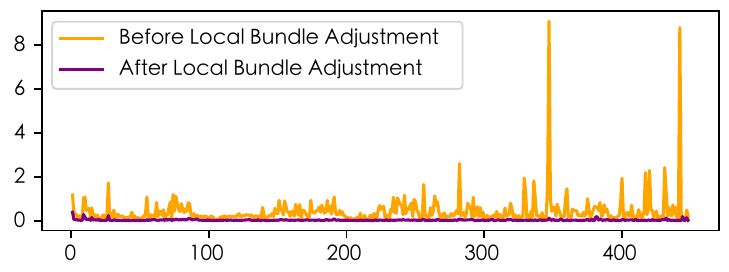}
	\end{subfigure}
	\begin{subfigure}[b]{0.23\textwidth}
		\includegraphics[width=\linewidth]{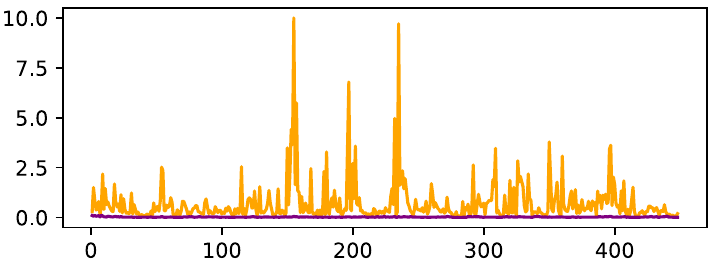}
		\label{fig:subfig1}
	\end{subfigure}
	\begin{subfigure}[b]{0.23\textwidth}
		\includegraphics[width=\linewidth]{figures/nsh_patio_p2p.pdf}
		\label{fig:subfig2}
	\end{subfigure}
	\vspace{-0.3cm}
	\caption{Distances between the 3D points and the corresponding LiDAR planes (indoor, outdoor1, outdoor2).}
	\label{fig:point_to_plane_dist}
\vspace{-0.3cm}
\end{figure}

The second experiment demonstrates the trend of the point-to-plane distances before and after each local bundle adjustment.
As Fig.~\ref{fig:point_to_plane_dist} shows, the distance converges to a minimal value after each iteration, indicating that 3D points can be aligned well to the appropriate positions in the point cloud map.

\subsection{Reprojection Error Analysis}
\begin{figure}[h]
	\begin{subfigure}{\textwidth}
		\includegraphics[width=0.35\textwidth]{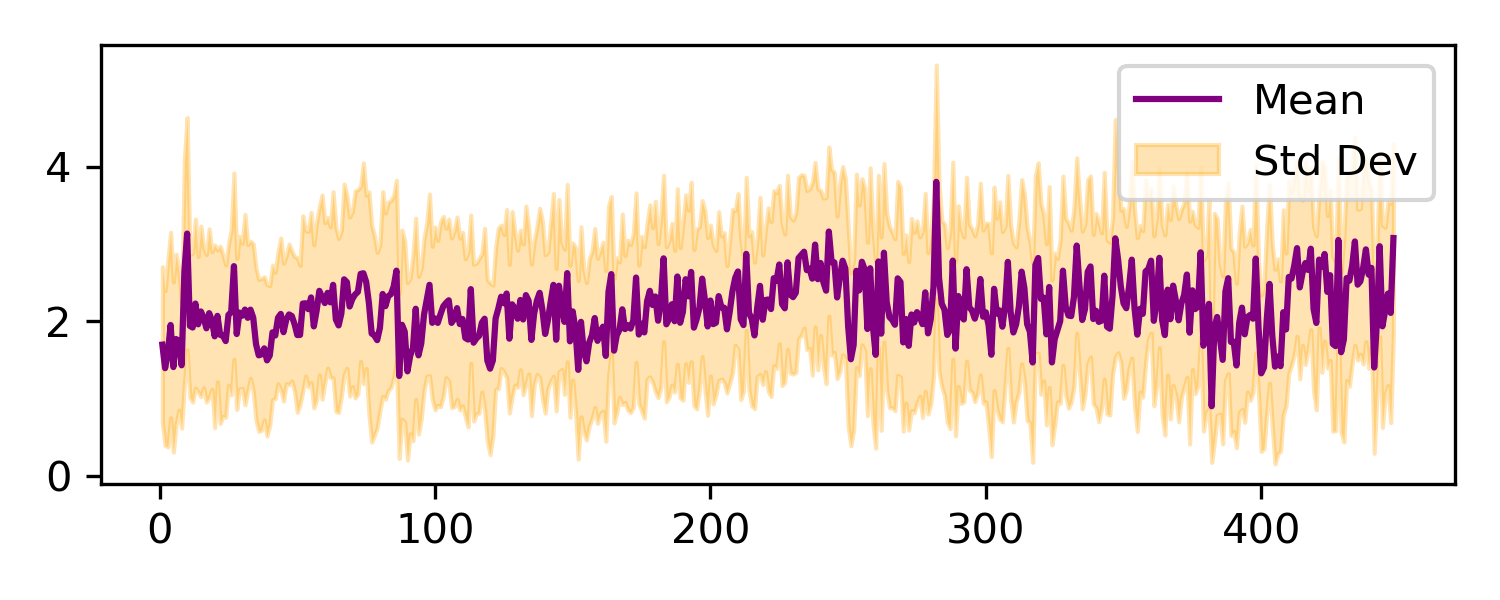}
	\end{subfigure}
	\begin{subfigure}[b]{0.23\textwidth}
		\includegraphics[width=\linewidth]{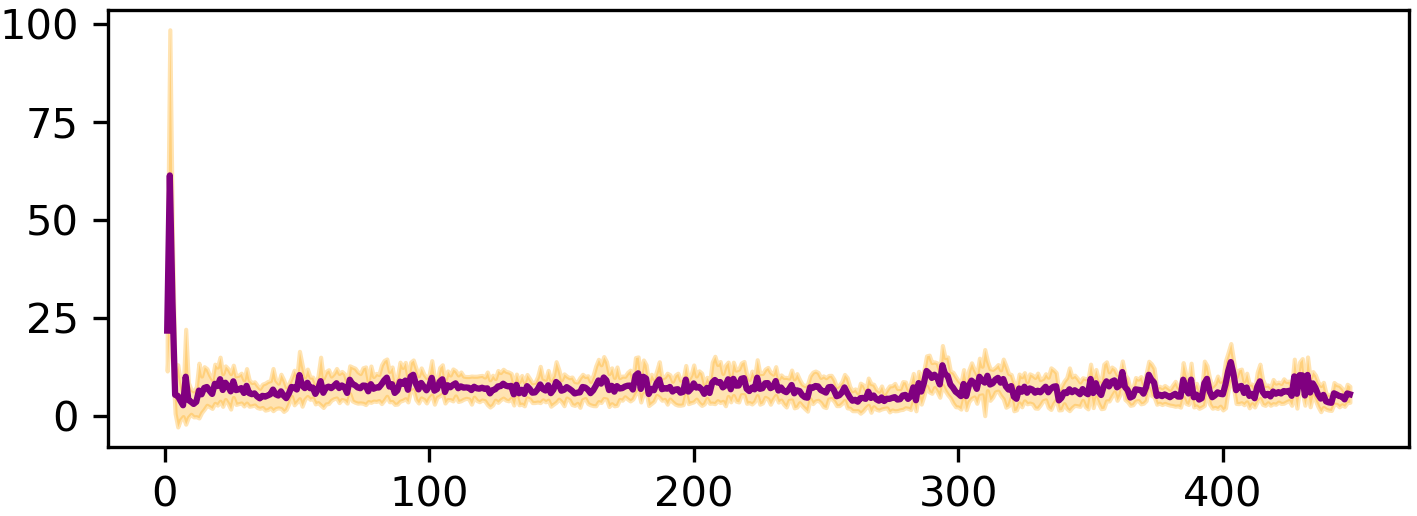}
		\label{fig:subfig1}
	\end{subfigure}
	\begin{subfigure}[b]{0.23\textwidth}
		\includegraphics[width=\linewidth]{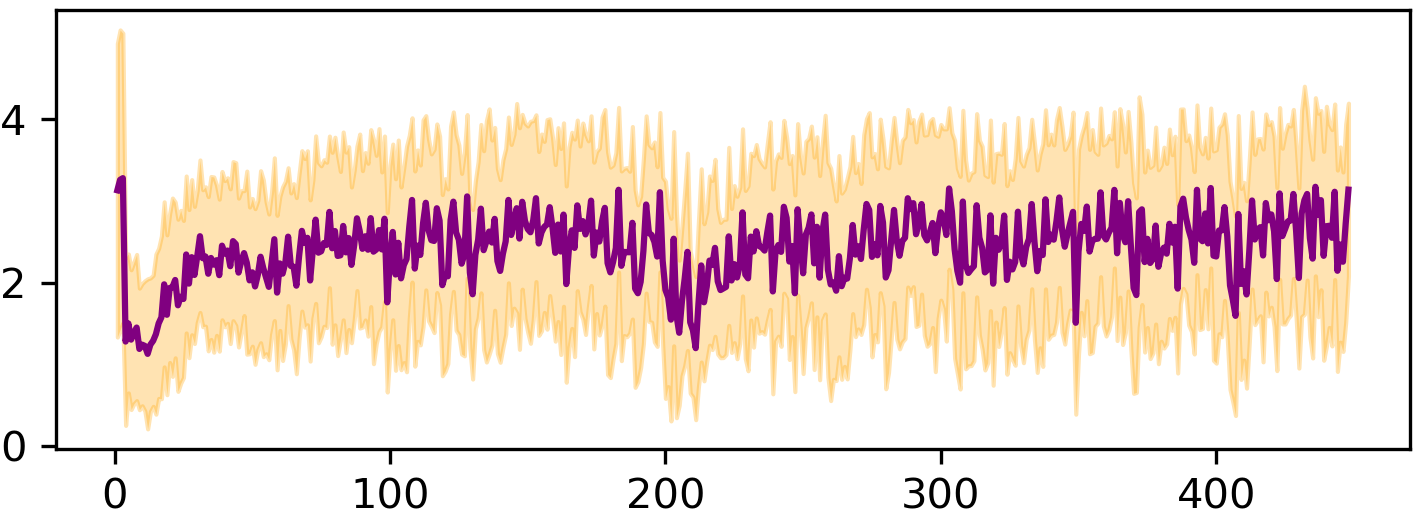}
		\label{fig:subfig2}
	\end{subfigure}
	\vspace{-0.3cm}
	\caption{Reprojection error (indoor, outdoor1, outdoor2).}
	\label{fig:reprojection_error}
\vspace{-0.6cm}
\end{figure}

This experiment demonstrates the variation of reprojection error. As Fig.~\ref{fig:reprojection_error} shows, the purple line represents the mean reprojection error, while the yellow area depicts the distribution of reprojection error covariance.
The reprojection error is calculated in the original image plane with an image size of 4032x3024. 
As a result, the reprojection error remains consistently at a small value, demonstrating that the method can maintain stability as the number of images increases.

\subsection{Initial Image Pose Error Test}
This experiment tested the accuracy requirements for the initial image pose. 
We captured the first image near the origin of the LiDAR map, so the initial position and orientation were set to 0. 
The final localization result of the initial image, as shown in the table \ref{tab:ini_err}, indicates that with the increase of images, even when providing a relatively large initial position error, the pose of the initial image can gradually converge to the correct result.
\begin{table}[h]
	\centering
	\caption{Localization result of the initial image}
	\begin{tabular}{ |c|c|c|c| } 
	\hline
	Dataset & Indoor Dataset & Outdoor Dataset1 & Outdoor Dataset2\\ 
	\hline
	$x$(m)& 0.0858479  & -0.0401451 & 1.68199 \\ 
	\hline
	$y$(m)& 0.320724 & 0.00475659 & -3.69749 \\ 
	\hline
	$z$(m)& 0.0244208 & -0.0773615 & 0.00797819 \\ 
	\hline
	$roll$(rad)& 0.020122 & 0.0208329 & -0.0113104 \\ 
	\hline
	$pitch$(rad) & -0.0124324 & -0.0515374 & -0.0204044 \\ 
	\hline
	$yall$(rad) & -0.00309135 & 0.0264514 & 0.441832  \\ 
	\hline
	\end{tabular}
	\label{tab:ini_err}
\vspace{-0.3cm}
\end{table}

\section{CONCLUSION}
This paper introduces a pipeline that aligns asynchronously acquired images with LiDAR point clouds to get localization results with an accurate scale. Through experiments on self-collected datasets, our approach demonstrates its effectiveness and stability in image accumulation. In addition, our pipeline enables flexible detail-level adjustments in the reconstruction area, making it suitable for collaborative large-scale scene reconstruction. In conclusion, our algorithm contributes to image localization and large-scale scene reconstruction tasks, offering valuable insights for future research in this field.

\addtolength{\textheight}{-12cm}  % This command serves to balance the column lengths
                                  % the text height of the last page by a suitable amount.
                                  % sure that you do not shorten the text height too much.

%%%%%%%%%%%%%%%%%%%%%%%%%%%%%%%%%%%%%%%%%%%%%%%%%%%%%%%%%%%%%%%%%%%%%%%%%%%%%%%%

\bibliographystyle{IEEEtran}
\bibliography{root}

\begin{thebibliography}{10}
\providecommand{\url}[1]{#1}
\csname url@samestyle\endcsname
\providecommand{\newblock}{\relax}
\providecommand{\bibinfo}[2]{#2}
\providecommand{\BIBentrySTDinterwordspacing}{\spaceskip=0pt\relax}
\providecommand{\BIBentryALTinterwordstretchfactor}{4}
\providecommand{\BIBentryALTinterwordspacing}{\spaceskip=\fontdimen2\font plus
\BIBentryALTinterwordstretchfactor\fontdimen3\font minus
  \fontdimen4\font\relax}
\providecommand{\BIBforeignlanguage}[2]{{%
\expandafter\ifx\csname l@#1\endcsname\relax
\typeout{** WARNING: IEEEtran.bst: No hyphenation pattern has been}%
\typeout{** loaded for the language `#1'. Using the pattern for}%
\typeout{** the default language instead.}%
\else
\language=\csname l@#1\endcsname
\fi
#2}}
\providecommand{\BIBdecl}{\relax}
\BIBdecl

\bibitem{schoenberger2016sfm}
J.~L. Sch\"{o}nberger and J.-M. Frahm, ``Structure-from-motion revisited,'' in
  \emph{Conference on Computer Vision and Pattern Recognition (CVPR)}, 2016.

\bibitem{schoenberger2016mvs}
J.~L. Sch\"{o}nberger, E.~Zheng, M.~Pollefeys, and J.-M. Frahm, ``Pixelwise
  view selection for unstructured multi-view stereo,'' in \emph{European
  Conference on Computer Vision (ECCV)}, 2016.

\bibitem{faster-lio}
C.~Bai, T.~Xiao, Y.~Chen, H.~Wang, and X.~Gao, ``Faster-lio: Lightweight
  tightly coupled lidar-inertial odometry using parallel sparse incremental
  voxels,'' 2023.

\bibitem{loam}
J.~Zhang and S.~Singh, ``Loam: Lidar odometry and mapping in real-time,''
  \emph{Robotics: Science and Systems}, 2014.

\bibitem{legoloam}
T.~Shan and B.~Englot, ``Lego-loam: Lightweight and ground-optimized lidar
  odometry and mapping on variable terrain,'' \emph{Intelligent Robots and
  Systems}, 2018.

\bibitem{YuePan2021MULLSVL}
Y.~Pan, P.~Xiao, Y.~He, Z.~Shao, and Z.~Li, ``Mulls: Versatile lidar slam via
  multi-metric linear least square,'' \emph{Cornell University - arXiv}, 2021.

\bibitem{sift}
T.~Lindeberg, ``Scale invariant feature transform,'' 2012.

\bibitem{bay2006surf}
H.~Bay, T.~Tuytelaars, and L.~Van~Gool, ``Surf: Speeded up robust features,''
  \emph{Lecture notes in computer science}, vol. 3951, pp. 404--417, 2006.

\bibitem{orb}
E.~Rublee, V.~Rabaud, K.~Konolige, and G.~Bradski, ``Orb: An efficient
  alternative to sift or surf,'' in \emph{2011 International conference on
  computer vision}.\hskip 1em plus 0.5em minus 0.4em\relax Ieee, 2011, pp.
  2564--2571.

\bibitem{forster2014svo}
C.~Forster, M.~Pizzoli, and D.~Scaramuzza, ``Svo: Fast semi-direct monocular
  visual odometry,'' in \emph{2014 IEEE international conference on robotics
  and automation (ICRA)}.\hskip 1em plus 0.5em minus 0.4em\relax IEEE, 2014,
  pp. 15--22.

\bibitem{engel2014lsd}
J.~Engel, T.~Sch{\"o}ps, and D.~Cremers, ``Lsd-slam: Large-scale direct
  monocular slam,'' in \emph{European conference on computer vision}.\hskip 1em
  plus 0.5em minus 0.4em\relax Springer, 2014, pp. 834--849.

\bibitem{MultipleVG}
R.~Hartley and A.~Zisserman, ``Multiple view geometry in computer vision,''
  2000.

\bibitem{icp1992}
P.~J. Besl and N.~D. McKay, ``Method for registration of 3-d shapes,'' in
  \emph{Sensor fusion IV: control paradigms and data structures}, vol.
  1611.\hskip 1em plus 0.5em minus 0.4em\relax Spie, 1992, pp. 586--606.

\bibitem{p3p}
X.-S. Gao, X.-R. Hou, J.~Tang, and H.-F. Cheng, ``Complete solution
  classification for the perspective-three-point problem,'' \emph{IEEE
  Transactions on Pattern Analysis and Machine Intelligence}, 2003.

\bibitem{upnp}
A.~Penate-Sanchez, J.~Andrade-Cetto, and F.~Moreno-Noguer, ``Exhaustive
  linearization for robust camera pose and focal length estimation,''
  \emph{IEEE Transactions on Pattern Analysis and Machine Intelligence}, 2013.

\bibitem{EPnP}
V.~Lepetit, F.~Moreno-Noguer, and P.~Fua, ``Epnp: An accurate o(n) solution to
  the pnp problem,'' \emph{International Journal of Computer Vision}, 2009.

\bibitem{p4p}
M.~Bujnak, Z.~Kukelova, and T.~Pajdla, ``A general solution to the p4p problem
  for camera with unknown focal length,'' \emph{Computer Vision and Pattern
  Recognition}, 2008.

\bibitem{ArnoldIrschara2009FromSP}
A.~Irschara, C.~Zach, J.-M. Frahm, and H.~Bischof, ``From structure-from-motion
  point clouds to fast location recognition,'' \emph{Computer Vision and
  Pattern Recognition}, 2009.

\bibitem{kaess2008isam}
M.~Kaess, A.~Ranganathan, and F.~Dellaert, ``isam: Incremental smoothing and
  mapping,'' \emph{IEEE Transactions on Robotics}, vol.~24, no.~6, pp.
  1365--1378, 2008.

\bibitem{okvis2013}
S.~Leutenegger, P.~Furgale, V.~Rabaud, M.~Chli, K.~Konolige, and R.~Siegwart,
  ``Keyframe-based visual-inertial slam using nonlinear optimization,'' 2013.

\bibitem{okvis2015}
S.~Leutenegger, S.~Lynen, M.~Bosse, R.~Siegwart, and P.~Furgale,
  ``Keyframe-based visual-inertial odometry using nonlinear optimization,''
  \emph{The International Journal of Robotics Research}, 2015.

\bibitem{zhang2015vloam}
J.~Zhang and S.~Singh, ``Visual-lidar odometry and mapping: Low-drift, robust,
  and fast,'' in \emph{2015 IEEE International Conference on Robotics and
  Automation (ICRA)}.\hskip 1em plus 0.5em minus 0.4em\relax IEEE, 2015, pp.
  2174--2181.

\bibitem{graeter2018limo}
J.~Graeter, A.~Wilczynski, and M.~Lauer, ``Limo: Lidar-monocular visual
  odometry,'' in \emph{2018 IEEE/RSJ international conference on intelligent
  robots and systems (IROS)}.\hskip 1em plus 0.5em minus 0.4em\relax IEEE,
  2018, pp. 7872--7879.

\end{thebibliography}

\end{document}